\documentclass[runningheads]{llncs}
\usepackage{graphicx}
\usepackage[caption=false]{subfig}
\usepackage{gensymb}
\usepackage{amsmath}
\usepackage{xcolor}
\usepackage{multirow}
\usepackage[linesnumbered,ruled]{algorithm2e}

\SetCommentSty{mycommfont}
\SetKwRepeat{Do}{do}{while}
\newcommand{\ineq}[1]{\footnotesize$#1$\normalsize}{}
\usepackage{threeparttable}

\newcommand{\mr}[1]{\textcolor{black}{#1}}

\begin{document}
%
\title{Thermal-Aware Compilation of Spiking Neural Networks to Neuromorphic Hardware}
\titlerunning{Thermal-aware SNN compilation}

\author{Twisha Titirsha\orcidID{00000-0002-2142-2283} \and
Anup Das\orcidID{0000-0002-5673-2636}}
%
\authorrunning{T. Titirsha et al.}
%
\institute{Drexel University, Philadelphia PA 19104, USA\\
\email{\{tt624,anup.das\}@drexel.edu}}
\maketitle              
\begin{abstract}
\mr{
Hardware implementation of neuromorphic computing can significantly improve performance and energy efficiency of machine learning tasks implemented with spiking neural networks (SNNs), making these hardware platforms particularly suitable for embedded systems and other energy-constrained environments.
} 
We observe that 
the long bitlines and wordlines in a crossbar of the hardware create significant current variations when propagating spikes through its synaptic elements, which are typically designed with non-volatile memory (NVM). Such current variations create a thermal gradient within each crossbar of the hardware, depending on the machine learning workload and the mapping of neurons and synapses of the workload to these crossbars. \mr{This thermal gradient becomes significant at scaled technology nodes and it increases the leakage power in the hardware leading to an increase in the energy consumption.} We propose a novel technique to map neurons and synapses of SNN-based machine learning workloads to neuromorphic hardware. We make two novel contributions. First, we formulate a detailed thermal model for a crossbar in a neuromorphic hardware incorporating workload dependency, where the temperature of each NVM-based synaptic cell is computed considering the thermal contributions from its neighboring cells. Second, we incorporate this thermal model in the mapping of neurons and synapses of SNN-based workloads using a hill-climbing heuristic. The objective is to reduce the thermal gradient in crossbars. We evaluate our neuron and synapse mapping technique using 10 machine learning workloads for a state-of-the-art neuromorphic hardware. \mr{We demonstrate an average 11.4K reduction in the average temperature of each crossbar in the hardware, leading to a 52\% reduction in the leakage power consumption (11\% lower total energy consumption) compared to a performance-oriented SNN mapping technique.}


\end{abstract}

\section{Introduction}\label{sec:introduction}
Spiking Neural Networks (SNNs) are 
machine learning models designed with spike-based computations and bio-inspired learning algorithms~\cite{maass1997networks}. Neurons communicate information using spikes via synapses.
SNNs are used to implement both supervised and unsupervised machine learning approaches. Our focus is on supervised approaches, where a machine learning model is first trained using training data, and then deployed for inference with in-field data.

\mr{Neuromorphic hardware such as TrueNorth \cite{truenorth}, Loihi~\cite{loihi}, and DYNAP-SE~\cite{dynapse} can significantly improve the energy efficiency of SNNs, thanks to their event-driven computations, efficient implementations of 
biological neurons using CMOS and FinFET technologies, and the use of Non-Volatile Memory (NVM) such as Phase-Change Memory (PCM)~\cite{Burr2017,hebe,mneme,palp,datacon}, Oxide-base Resistive RAM (OxRRAM)~\cite{Mallik2017}, and Spin-Transfer Torque Magnetic or Spin-Orbit-Torque RAM (STT- and SoT-MRAM)~\cite{ramasubramanian2014spindle} for high density synaptic storage.
Therefore, neuromorphic hardware can be used to implement machine learning tasks on power-constrained environments such as embedded systems, and sensor and edge devices of the Internet-of-Things (IoT)~\cite{gubbi2013internet}.}

\begin{figure}[h!]
	\centering
	\vspace{-5pt}
	\centerline{\includegraphics[width=0.99\columnwidth]{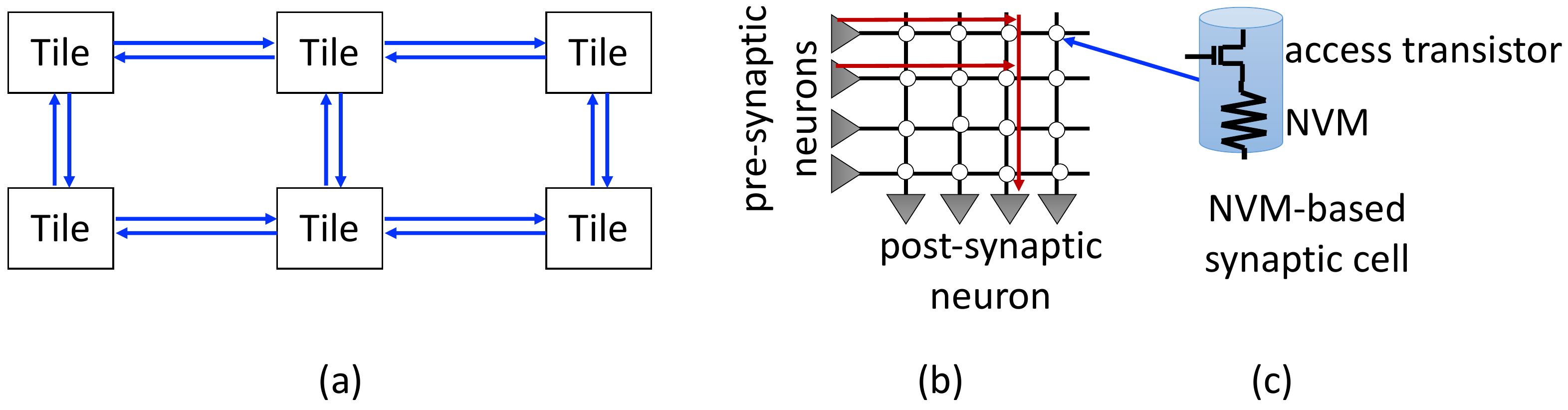}}
	\vspace{-10pt}
	\caption{(a) Tile-based neuromorphic hardware. (b) A crossbar of a neuromorphic tile. (c) An NVM-based synaptic cell consisting of an access transistor and an NVM.}
	\label{fig:architecture}
\end{figure}

A neuromorphic hardware is implemented as a tile-based architecture~\cite{catthoor2018very} with a shared interconnect in the form of Networks-on-Chip (NoC) or Segmented Bus~\cite{balaji2019exploration} (see Figure~\ref{fig:architecture}a). A tile in a neuromorphic hardware is designed as a crossbar, which is an organization of top electrodes (wordlines) and bottom electrodes (bitlines), with NVM-based synaptic elements at their intersections (Figure~\ref{fig:architecture}b). A synaptic element is connected to a bitline and a wordline using an access transistor (Figure~\ref{fig:architecture}c). Within a crossbar, the pre-synaptic neurons are mapped on the wordlines, while the post-synpatic neurons are mapped along the bitlines. An \ineq{n\times n} crossbar has \ineq{n} pre-synaptic neurons, \ineq{n} post-synaptic neurons, and \ineq{n^2} NVM cells. A pre-synaptic neuron's spike voltage from a wordline is multiplied with the conductance of the NVM to generate a current. Currents from multiple wordlines are integrated on a bitline, implementing forward propagation of neuron excitation. This is illustrated in Figure~\ref{fig:architecture}b.

\begin{figure}[h!]
	\centering
	\centerline{\includegraphics[width=0.5\columnwidth]{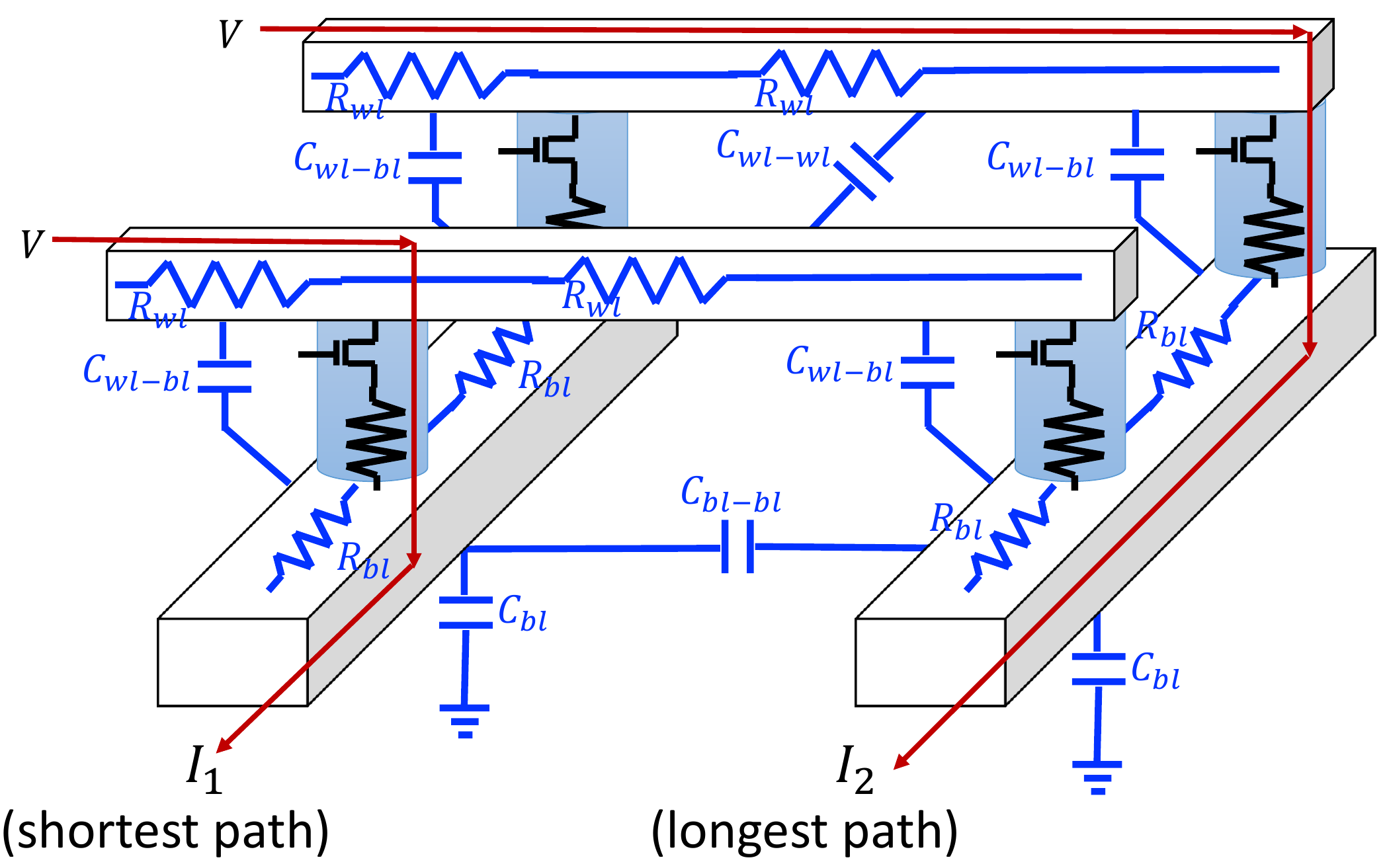}}
	\vspace{-10pt}
	\caption{Parasitc components on the bitlines and wordlines in a crossbar.}
	\label{fig:parasitics}
\end{figure}

\mr{We investigate the internal architecture of a crossbar and observe that the bitlines and wordlines of a crossbar consist of parasitic elements, which consist of capacitance and resistance of the metal interconnect as shown in Figure~\ref{fig:parasitics}.} These parasitic elements create variation in current propagating along different paths in the crossbar. The figure illustrates the shortest and the longest current paths in a crossbar, where the length of a path is measured in terms of the number of parasitic components that are present on the path. Current differences create variation in access speed of the different synaptic elements in the crossbar~\cite{fouda2017modeling,woo2018resistive,twisha_endurance}. A conservative design practice is to use a common spike voltage to obtain the required access speed of the synaptic element on the longest current path. 

\begin{figure}[h!]%
    \centering
    \subfloat[Current variation for PCM access operations in a 128x128 crossbar.]{{\includegraphics[width=5.6cm]{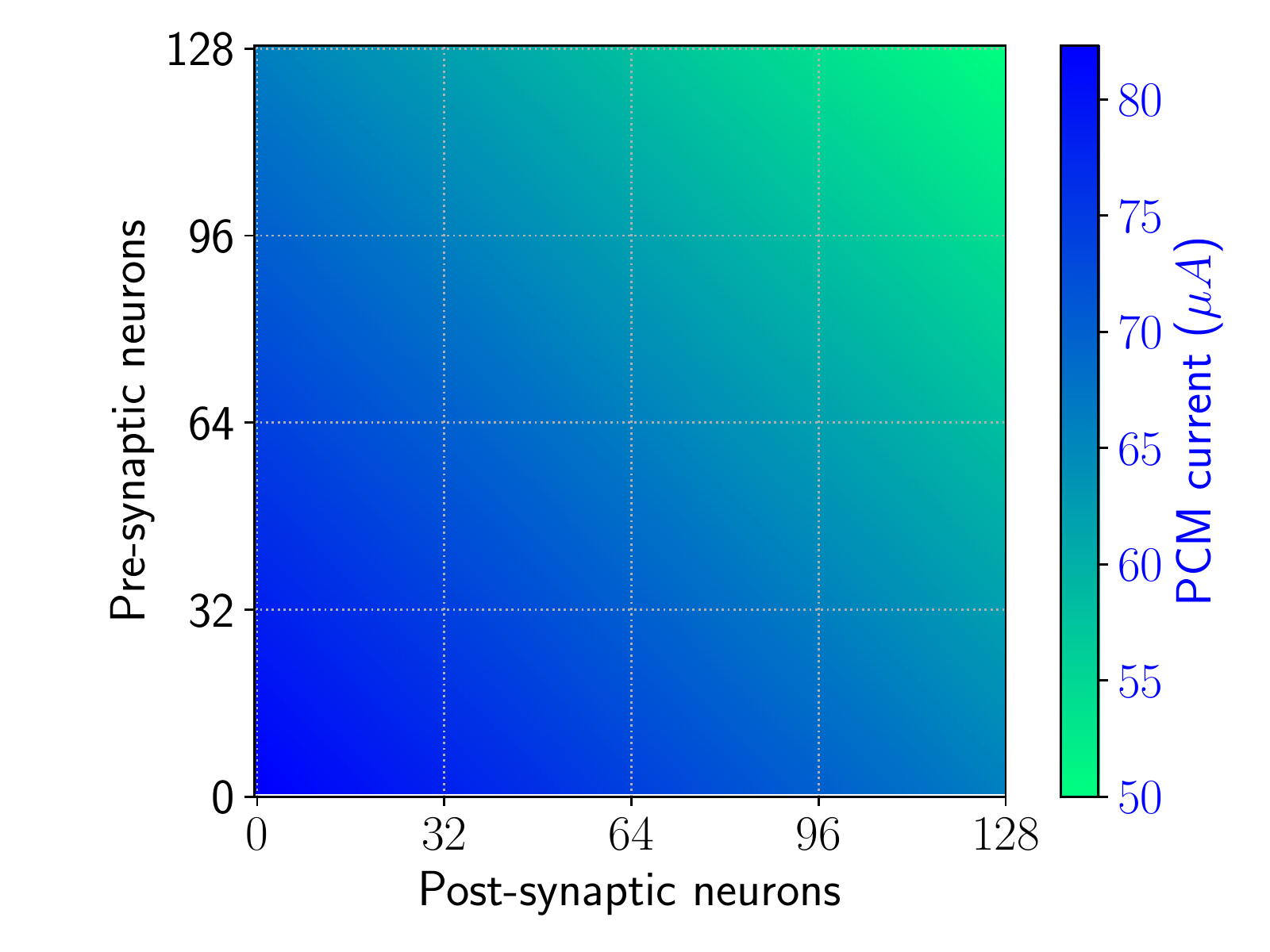} }}%
    \quad
    \subfloat[Temperature gradient in a 128x128 crossbar.]{{\includegraphics[width=5.6cm]{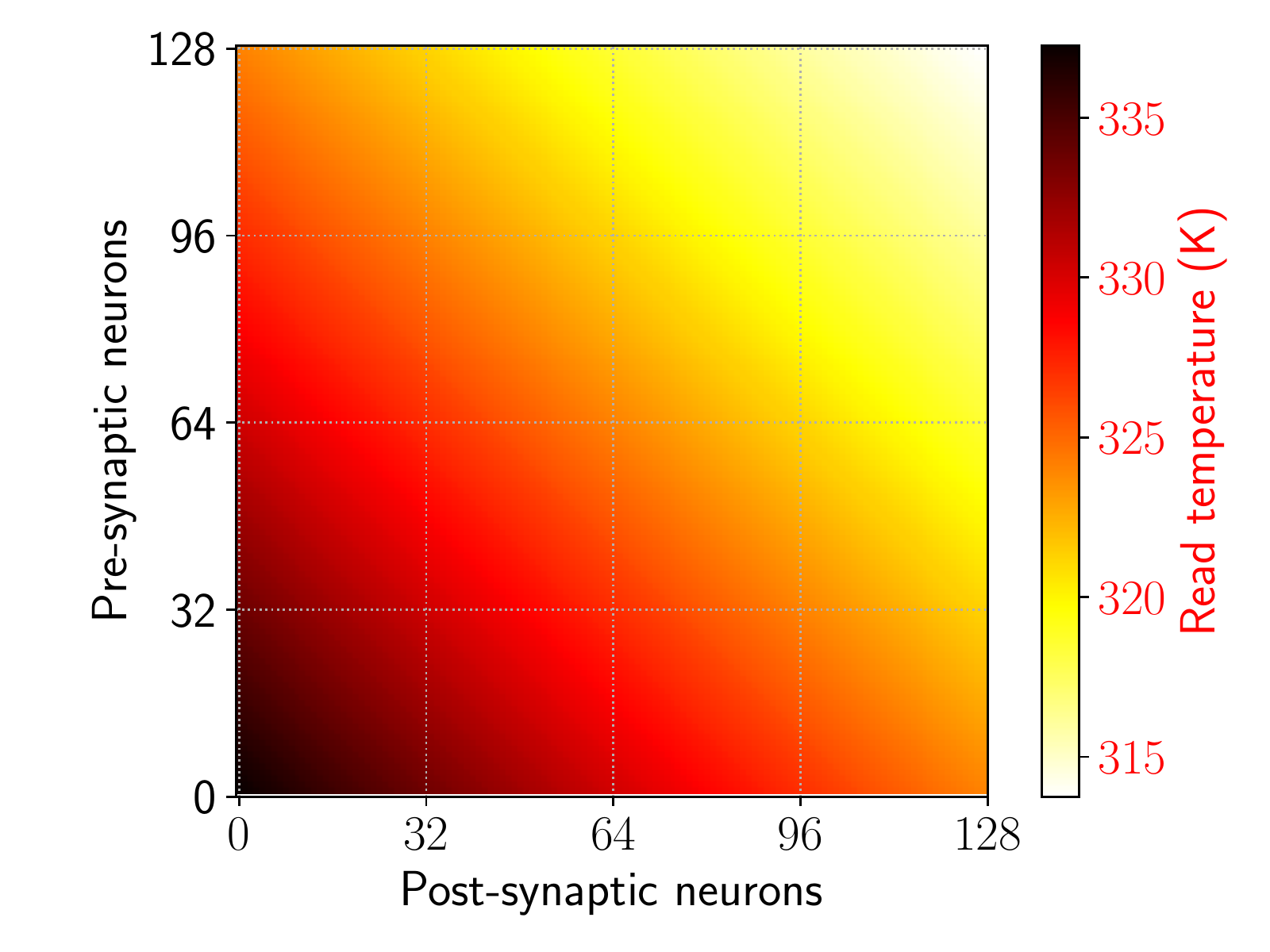} }}%
    \caption{Current variation and temperature gradient in a 128x128 crossbar at \ineq{65nm} process node with \ineq{T_{amb} = 298K}. The PCM crystallization point is 360K.}%
    \label{fig:thermal_gradient}%
\end{figure}

We argue that this conservative approach creates current differences in a crossbar, leading to a wide thermal gradient. Figure~\ref{fig:thermal_gradient} illustrates the current and thermal variations in a 128x128 PCM crossbar at 65nm technology node. Accessing the synaptic cells on shorter current paths (bottom left corner of Figure~\ref{fig:thermal_gradient}b) generate higher temperatures than those on longer current paths (top right corner). \mr{Due to the exponential dependency of leakage current on temperature~\cite{liu2007accurate}, the leakage current through cells with higher temperature is much higher than the current through cells with lower temperature. So, frequently accessing the cells on shorter current paths when executing a workload can lead to higher leakage power consumption in the crossbar.}

\mr{Existing techniques to map neurons and synapses of SNNs to neuromorphic hardware have mostly focused on improving performance and circuit aging~\cite{psopart,das2018dataflow,balaji2019frameworkISVLSI,spinemap,frameworkCAL,dfsynthesizer,reneu,NeuromorphicLR,balaji2020run,balaji2020ESL}. 
These techniques do not consider the thermal gradient in a crossbar and therefore, they can increase the leakage power significantly.
We build the case for one such mapping technique -- SpiNeMap~\cite{spinemap}. The leakage energy using this technique constitute between 20\% to 30\% of the total energy consumption for the typical machine learning workloads (see Section~\ref{sec:evaluation}), where the total energy of a neuromorphic hardware includes the energy to generate spikes, the energy to communicate spikes, and the leakage energy. Therefore, reducing the leakage power (which we demonstrate in this work) will lead to a significant reduction of the total energy consumption.
}

\mr{
Our \textbf{goal} is to minimize the leakage power consumption. We achieve this goal by lowering the average temperature of each crossbar using the proposed mapping technique. To this end, we make the following two key contributions.}
\begin{itemize}
    \item \textbf{Contribution 1:} We propose a new comprehensive thermal model of a crossbar designed with phase-change memory (PCM). Our model incorporates 1) workload dependency, i.e., the temperature obtained in processing spike trains from a given SNN-based machine learning workload, and 2) spatial thermal dependencies, i.e., the temperature contributions from the neighboring cells based on their synaptic excitation in the workload.
    \item \textbf{Contribution 2:} We propose a novel neuron and synapse mapping approach incorporating the thermal model using a hill climbing heuristic. The objective of the heuristic is to allocate the neurons and synapses of an SNN to the crossbars of the hardware such that the maximum average temperature of all crossbars is minimized, which lowers its leakage power consumption.
\end{itemize}

We evaluate the proposed technique with 10 machine learning applications from three most commonly-used neural network topology -- convolution neural network (CNN), multilayer perceptron (MLP), and recurrent neural network (RNN). Evaluation for DYNAP-SE~\cite{dynapse}, a state-of-the-art neuromorphic hardware demonstrates the reduction of temperature, leading to a significant reduction in the leakage current.

\section{Workload-dependant Thermal Model of Crossbars}\label{sec:crossbars}
In this section, we develop a workload-dependent thermal model of crossbars in a neuromorphic hardware, considering PCM-based synaptic elements. We start by reviewing the internals of a PCM device. The proposed thermal model can be generalized to other NVMs such as OxRRAM and SOT-/STT-MRAM exploiting their specific structures.

\begin{figure}[h!]
	\begin{center}
		\vspace{-10pt}
		\includegraphics[width=0.79\columnwidth]{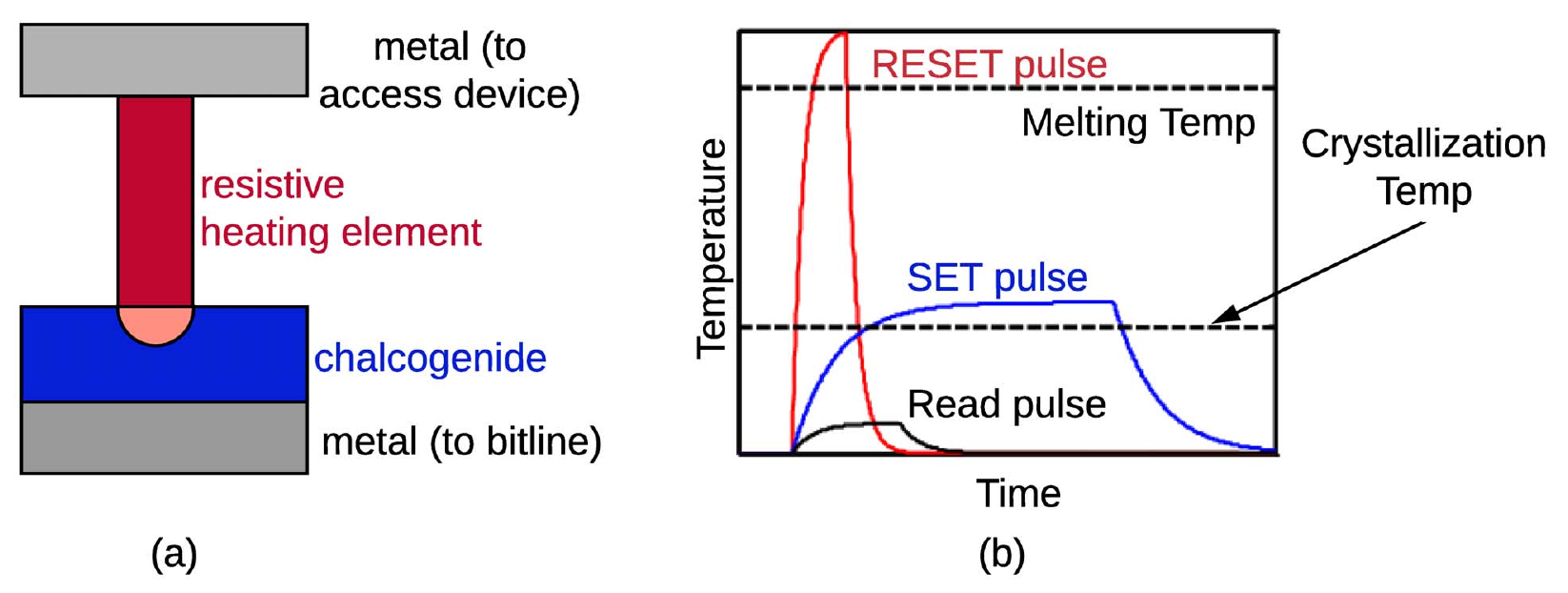}
		\vspace{-10pt}
		\caption{(a) A PCM cell and (b) Current needed to operate a PCM cell.}
		\vspace{-10pt}
		\label{fig:pcm_memory_cell_integration}
		\vspace{-10pt}
	\end{center}
\end{figure}

Figure \ref{fig:pcm_memory_cell_integration}(a) illustrates how a chalcogenide semiconductor alloy is used to build a PCM cell.
{
The amorphous phase (logic `0') in this alloy has higher resistance than the crystalline phase (logic `1').
}
Ge${}_2$Sb${}_2$Te${}_5$ (GST) is the most commonly used alloy for PCM~\cite{wong2010phase} due to its high amorphous-to-crystalline resistance ratio, fast switching between phases, and high endurance.
However, other chalcogenide alloys are also explored due to their better data retention properties~\cite{morikawa2007doped}.
Phase changes in a PCM cell are induced by injecting current into the resistor-chalcogenide junction and heating the chalcogenide alloy. 

{
Figure \ref{fig:pcm_memory_cell_integration} (b) shows the different current profiles needed to program and read in a PCM device.
}
To RESET a PCM cell, a high power pulse of short duration is applied and quickly terminated. This first raises the temperature of the chalcogenide alloy to 650$\degree$C, above its melting point. The melted alloy subsequently cools extremely quickly, locking into an amorphous phase.
To SET a PCM cell, the chalcogenide alloy is heated above its crystallization temperature, but below its melting point for a sufficient amount of time. 
Finally, to read the content (i.e., know the phase) of a PCM cell, a small electrical pulse is applied that is sufficiently low so as not to induce phase change in the PCM cell. 
We focus on PCM read for the inference of supervised machine learning approaches

\begin{figure}[h!]%
    \centering
    \subfloat[Thermal interactions in a PCM-based crossbar.]{{\includegraphics[width=5.6cm]{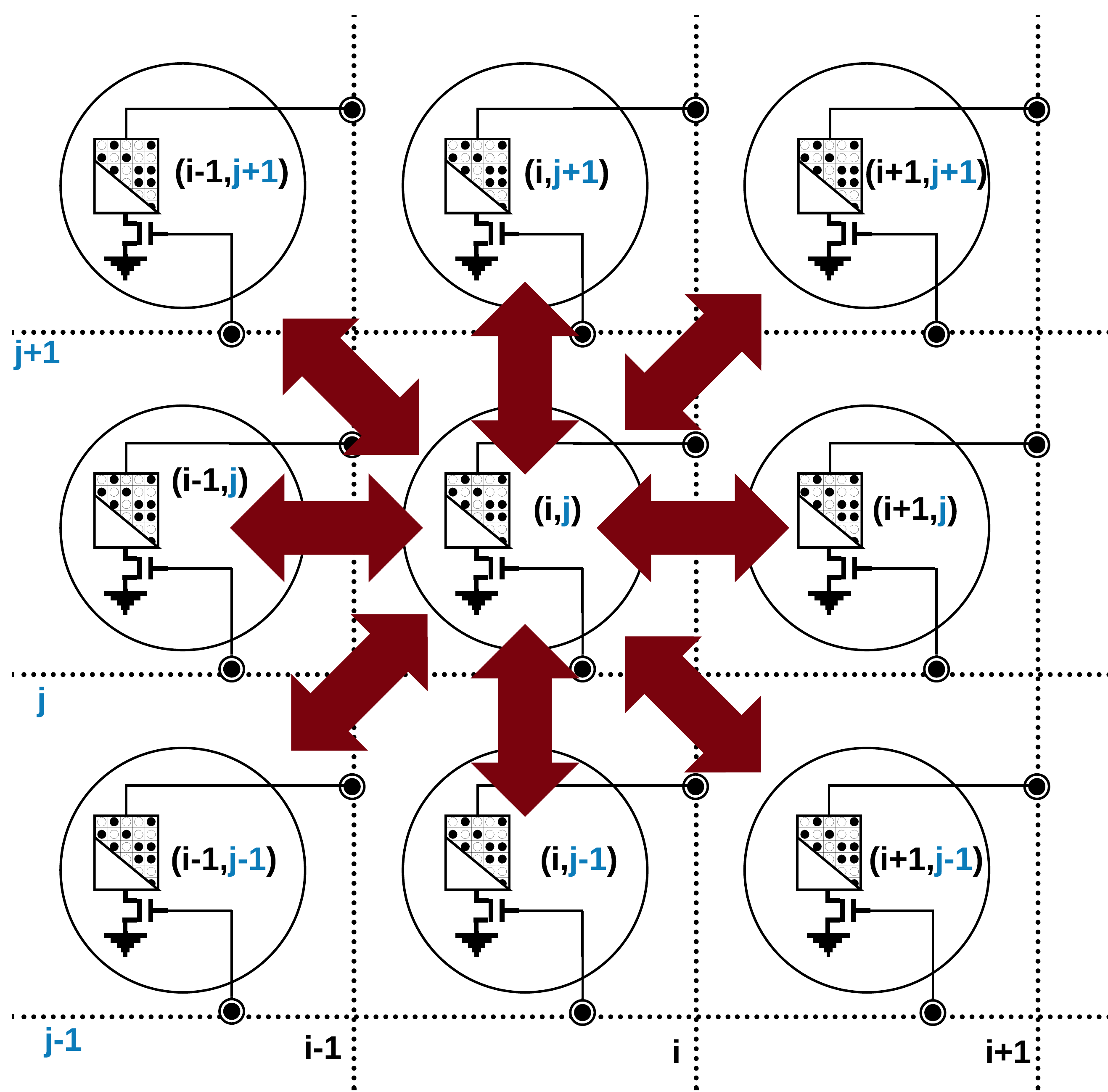} }}%
    \quad
    \subfloat[Iterative thermal computations.]{{\includegraphics[width=5.6cm]{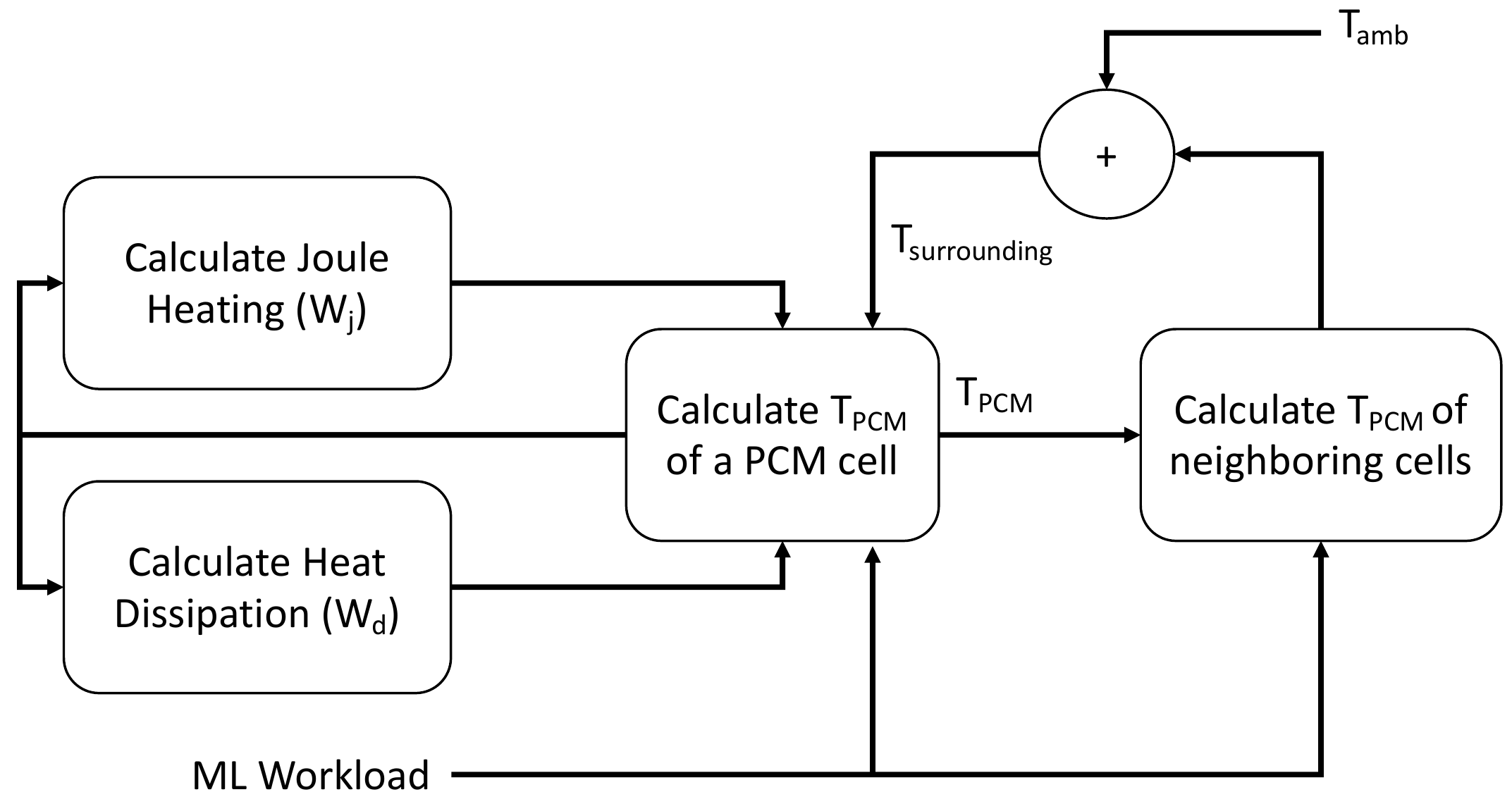} }}%
    \caption{Building thermal model of a PCM-based crossbar.}%
    \label{fig:interactions}%
\end{figure}

Many prior works have developed thermal models for PCM devices~\cite{warren2008compact,chen2009compact,le2016evidence}. However, these models are developed for individual PCM cell considering the effect of crystallization and amorphization (synaptic weight updates in the context of machine learning). In other words, these models have the following two key limitations for their use in the context of neuromorphic computing. First, they do not consider spatial dependencies, i.e., the thermal contributions from neighboring PCM cells considering their utilization in a machine learning workload. Second, the thermal impact due to PCM reads (as required for machine learning inference) is not modeled. Figure~\ref{fig:interactions}a shows the thermal interactions in a crossbar. When a cell is accessed repeatedly within a short time window, there remains very little scope for heat inside the cell to be dissipated. As a result temperature keeps rising on every access, building on the undissipated components, and dissipating the heat to its neighboring cells, raising their temperature.

Figure~\ref{fig:interactions}b shows the proposed iterative approach of computing the temperature of a crossbar. The model computes the temperature of a single PCM cell incorporating 1) thermal contributions from its neighbors and 2) its activation within a workload. 

The temperature of a single PCM cell is computed using Joule heating, $W_j$ and heat dissipation, $W_d$ which is given by the following equation~\cite{Liao2008},
\begin{equation}
\label{eq:TPCM}
T_{PCM}=\int \frac{W_j-W_d}{C\times V}dt
\end{equation}
where C and V are heat capacity of GST and volume of the active region of the cell respectively. The heat generation in the PCM cell is given by,
\begin{equation}
\label{eq:joule heating}
W_j=I_{PCM}^2\times R_{PCM}
\end{equation}
where $I_{PCM}$ is the current through the PCM cell and $R_{PCM}$ is the effective resistance of the cell. We use $R_{PCM}$ = 10K$\Omega$ in the low resistance (SET) state and 200K$\Omega$ in the high resistance (RESET) state. A part of this generated heat is dissipated to the surrounding and this heat dissipation is given by the Equation~\cite{Liao2007},
\begin{equation}
\label{eq:heat dissipation}
W_d=-k\sum\Delta T
\end{equation}
where $\Delta T$ represents the temperature dispersion around the active region and expressed as, 
\begin{equation}
\label{eq:delta_T}
    \Delta T=\frac{\partial T_{PCM}}{\partial x}+\frac{\partial T_{PCM}}{\partial y}+ \frac{\partial T_{PCM}}{\partial z}
\end{equation}
For simplicity we assume that the heat is mainly dispersed along the thickness of the cell and the temperature outside the dispersion region is close to the temperature surrounding the cell. Therefore, Equation~\ref{eq:heat dissipation} can be written as~\cite{Kwong2008},~\cite{Wei2012},
\begin{equation}
\label{eq:Wd}
    W_d=\frac{kV}{l^2}(T_{PCM}-T_{surrounding})
\end{equation}
where $l$ is the thickness of the GST material and $k$ is the thermal conductivity. Substitution of Equations~\ref{eq:joule heating} and ~\ref{eq:Wd} in Equation ~\ref{eq:TPCM} yields,
\begin{equation}
    \frac{dT_{PCM}}{dt}=\frac{W_j-W_d}{C\times V}
\end{equation}
Solving this ODE gives,
\begin{equation}
    T_{PCM}=\frac{I_{PCM}^2R_{PCM}l^2}{kV}-C_1 exp\left(-\frac{kt}{l^2C}\right)+T_{surrounding}
\end{equation}
Initially the PCM cell's temperature is assumed to be the same as its surrounding temperature. This boundary condition is used to determine the constant $C_1$. Finally the cell temperature is modeled as, 
\begin{equation}
\label{eq:final_TPCM}
    T_{PCM}=\frac{I_{PCM}^2R_{PCM}l^2}{kV}-\left[1-exp\left(-\frac{kt}{l^2C}\right)\right]+T_{surrounding}
\end{equation}
The surrounding temperature $T_{surrounding}$ is computed as
\begin{equation}
    \label{eq:surrounding}
    T_{surrounding} = T_{amb} + \sum_j k\cdot T_{PCM_j} / D_j
\end{equation}
where $D_j$ is the thermal distance of the PCM cell from its neighboring cell $j$, $T_{PCM_j}$ is the temperature of the neighboring cell, and $T_{amb}$ is the ambient temperature of the neuromorphic hardware.

Equations~\ref{eq:final_TPCM} and \ref{eq:surrounding} combine the following effects --- 1) temporal thermal effect of accessing a PCM cell in a machine learning workload, 2) the spatial thermal contributions from the neighboring cell based on their activation.

Finally, we use the PCM temperature $T_{PCM}$ to compute the leakage current through the access transistor of the PCM cell using Equation~\ref{eq:leakage}, where the fitting parameters $A$ and $\eta$, and the nominal parameters $I_{nominal}$ and $T_{nominal}$ are obtained using~\cite{liu2007accurate,shafik2015adaptive,das2015hardware,das2014communication,das2013communication,balaji2018power,das2018energy}.
\begin{equation}
    \label{eq:leakage}
    I_{leakage} \approx A\cdot I_{nominal}\left(T_{PCM}-T_{nominal}\right){}^{\eta}
\end{equation}

\section{Proposed Neuron and Synapse Mapping Technique}\label{sec:mapping}
\begin{figure}[h!]
	\begin{center}
		\vspace{-10pt}
		\includegraphics[width=0.79\columnwidth]{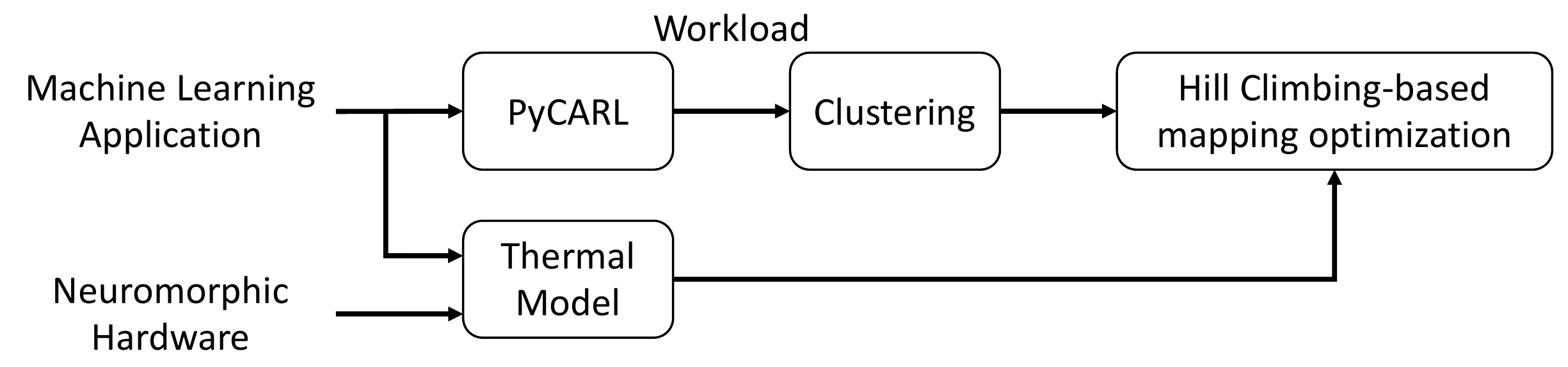}
		\vspace{-10pt}
		\caption{Overview of the proposed technique.}
		\vspace{-10pt}
		\label{fig:mapping}
		\vspace{-10pt}
	\end{center}
\end{figure}

Figure~\ref{fig:mapping} shows an overview of the proposed neuron and synapse mapping approach. A machine learning application is first simulated using PyCARL~\cite{pycarl}, a framework for simulating SNN-based applications. PyCARL internally uses the CARLsim~\cite{carlsim} simulator to extract the precise spike times on every synaptic element in the SNN for representative training data. These spike times, together with the neuron and synapse information constitute the SNN workload for the machine learning application.

Next, the SNN workload is clustered using a greedy clustering approach, roughly based on the Kernighan-Lin Graph Partitioning algorithm of SpiNe\-Map~\cite{kernighan1970efficient}. 
\mr{
Each cluster is a collection of pre- and post-synaptic neurons, synapses connecting these neurons, and the spike times on these synapses.
From the mapping perspective, each cluster maps to a crossbar in the hardware, while the inter-cluster communication channels are mapped on the shared interconnect of the hardware. Therefore, the clustering technique ensures that the neurons and synapses of a cluster can fit onto the resources of the crossbar.
PyCARL clusters an SNN to minimize the inter-cluster communication. This reduces the spike congestion on the shared interconnect which improves application latency.
}

The final step in our approach is the cluster mapping to the hardware. To describe this step, let \ineq{{G} ({C,S})} be the machine learning workload with set \ineq{{C}} of clusters and a set \ineq{{S}} of connections between the clusters. The workload is to be executed on the hardware \ineq{H(\mathcal{T},L)} with a set \ineq{\mathcal{T}} of tiles (each tile has one crossbar) and a set \ineq{L} of links between the tiles. The mapping of the application \ineq{G} to the hardware \ineq{H}, \ineq{\mathcal{M} = \{m_{x,y}\}} is defined as
\begin{equation}
    \label{eq:mapping_rep}
    \footnotesize m_{x,y} = \begin{cases}
    1 & \text{if cluster } {c}_x \in {C} \text{ is mapped to tile } {t}_y\in{T}\\
    0 & \text{otherwise}
    \end{cases}
\end{equation}

\vspace{-10pt}

\begin{algorithm}[h]
	\scriptsize{
	    \KwIn{$G,H$}
	    \KwOut{$\mathcal{M}$}
	    \For{$i$ in $MaxIter$}{
	        $\mathcal{M}_\text{init}$ = allocate clusters to crossbars randomly\;
	        $T_\text{init}$ = \texttt{CalculateAvgTemperature}($\mathcal{M}_\text{init}$)\;
	        \Do{$T_\text{min} < T_\text{init}$}{
	            \For{$c_x\in C$}{
	                \For{$t_y\in \mathcal{T}$}{
	                    $\mathcal{M}_y = \mathcal{M}_\text{init}\big| m_{x,z} = \begin{cases}
                        1 & \text{if } z = y\\
                        0 & \text{otherwise}
                        \end{cases}$ \tcc{Move $c_x$ to tile $t_y$ and generate the new mapping $\mathcal{M}_{y}$.}
                        $T_y$ = \texttt{CalculateAvgTemperature}($\mathcal{M}_y$)\;
	                }
	                $x_\text{idx}$ = \texttt{argmin} $\{T_y\big|y\in 1,2,\cdots,|\mathcal{T}|\}$\tcc{Find the index of the mapping with the minimum temperature.}
	                \If{$T_y < T_\text{min}$}{
	                    $T_\text{min} = T_y$ and $\mathcal{M}_\text{min} = \mathcal{M}_y$\tcc{Update the mapping if the average temperature reduces.}
	                }
	            }
	        }
	    }
	    Return $\mathcal{M}_\text{init}$
	}
	\caption{\small Generate neuron and synapse mapping \ineq{\mathcal{M}} to minimize the average temperature of crossbars.}
	\label{alg:min_temp}
\end{algorithm}

\vspace{-10pt}

Algorithm~\ref{alg:min_temp} provides the pseudo-code of the hill-climbing based average temperature minimization algorithm. The algorithm takes the clustered application $G$ and the neuromorphic hardware $H$ as input. The algorithm returns the mapping of $G$ to $H$, which minimizes the average temperature of the crossbars. The algorithm is iterated for \ineq{MaxIter} iterations (outer loop lines 1-16). For each iteration of the outer loop, the algorithm generates a random allocation of the clusters to the tiles (line 2) and calculate the average temperature (line 3). The routine \texttt{CalculateAvgTemperature} calculates the temperature of each crossbar for a mapping \ineq{\mathcal{M}} using the iterative approach of Figure~\ref{fig:interactions}b, specifically utilizing Equations~\ref{eq:final_TPCM} \& \ref{eq:surrounding}, and return the maximum average temperature of all crossbars in the neuromorphic hardware. 

At each iteration of the Algorithm~\ref{alg:min_temp}, a cluster is moved to one of the tiles (line 7), computing the average temperature of this new mapping (line 8). The one mapping that leads to reduction of the average temperature is retained as the new mapping (lines 10-13) and the process is repeated for the next cluster (5-14). Once every cluster is analyzed, the iteration is repeated (lines 4-15) to check if the clusters can be remapped again to reduce the average temperature.
The user-defined parameter \ineq{MaxIter} governs the convergence of the algorithm.

\textbf{Algorithm Complexity:} The complexity of Algorithm~\ref{alg:min_temp} is calculated as follows. Let the inner loop (lines 4-15) be executed \ineq{\zeta} times on average. At each of these iterations, the algorithm performs \ineq{|C|\times |\mathcal{T}|} operations. Therefore, the complexity of Algorithm~\ref{alg:min_temp} is \ineq{O(MaxIter\times\zeta\times|C|\times|\mathcal{T}|)}.

\section{Evaluation}\label{sec:evaluation}
\subsection{Evaluated Applications}
We evaluated 10 machine learning applications that are representative of three most commonly used neural network classes --- convolutional neural network (CNN), multi-layer perceptron (MLP), and recurrent neural network (RNN).
Table~\ref{tab:apps} summarizes the topology, the number of neurons and synapses of these applications, and their baseline accuracy.

\vspace{-10pt}

\begin{table}[h!]
	\renewcommand{\arraystretch}{0.8}
	\setlength{\tabcolsep}{2pt}
	\caption{Applications used to evaluate the proposed technique.}
	\label{tab:apps}
	\vspace{-5pt}
	\centering
	\begin{threeparttable}
	{\fontsize{6}{10}\selectfont
		\begin{tabular}{cc|ccl|c}
			\hline
			\textbf{Class} & \textbf{Applications} & \textbf{Synapses} & \textbf{Neurons} & \textbf{Topology} & \textbf{Accuracy}\\
			\hline
			\multirow{4}{*}{CNN} & LeNet~\cite{lenet} & 282,936 & 20,602 & CNN & 85.1\%\\
			& AlexNet~\cite{alexnet} & 38,730,222 & 230,443 & CNN & 90.7\%\\
			& VGG16~\cite{vgg16} & 99,080,704 & 554,059 & CNN & 69.8 \%\\
			& HeartClass~\cite{HeartClassJolpe,das2018heartbeat} & 1,049,249 & 153,730 & CNN & 63.7\%\\
			\hline
			\multirow{3}{*}{MLP} & DigitRecogMLP & 79,400 & 884 & FeedForward (784, 100, 10) & 91.6\%\\
			& EdgeDet \cite{carlsim} & 114,057 &  6,120 & FeedForward (4096, 1024, 1024, 1024) & 100\%\\
			& ImgSmooth \cite{carlsim} & 9,025 & 4,096 & FeedForward (4096, 1024) & 100\%\\
			\hline
 			\multirow{3}{*}{RNN} & HeartEstm \cite{HeartEstmNN} & 66,406 & 166 & Recurrent Reservoir & 100\%\\
 			& VisualPursuit \cite{Kashyap2018} & 163,880 & 205 & Recurrent Reservoir & 47.3\%\\
 			& R-DigitRecog \cite{Diehl2015} & 11,442 & 567 & Recurrent Reservoir & 83.6\%\\
			\hline
	\end{tabular}}
	\end{threeparttable}
	\vspace{-10pt}
\end{table}

\subsection{Hardware Models}
We model the DYNAP-SE neuromorphic hardware~\cite{dynapse} with the following configurations.

\begin{itemize}
    \item A tiled array of 4 tiles, each with a 128x128 crossbar. There are 65,536 memristors per crossbar.
    \item Spikes are digitized and communicated between cores through a mesh routing network using the Address Event Representation (AER) protocol.
    \item Each synaptic element is a PCM-based memristor. 
\end{itemize}

Table \ref{tab:hw_parameters} reports the hardware parameters of DYNAP-SE.

\vspace{-10pt}

\begin{table}[h!]
    \caption{Major simulation parameters extracted from \cite{dynapse}.}
	\label{tab:hw_parameters}
	\vspace{-10pt}
	\centering
	{\fontsize{6}{10}\selectfont
		\begin{tabular}{lp{5cm}}
			\hline
			Neuron technology & 32nm FD-SOI\\
			\hline
			Synapse technology & PCM\\
			\hline
			Supply voltage & 1.0V\\
			\hline
			Energy per spike & 50pJ at 30Hz spike frequency\\
			\hline
			Energy per routing & 147pJ\\
			\hline
			Switch bandwidth & 1.8G. Events/s\\
			\hline
	\end{tabular}}
	\vspace{-10pt}
\end{table}

\subsection{Evaluated Techniques}
We evaluate the following two approaches.
\begin{itemize}
    \item \textbf{SpiNeMap~\cite{spinemap}:} This is a performance-oriented approach to map SNN-based applications to neuromorphic hardware. This approach first generates clusters of neurons and synapses, where each cluster can fit on to the resources of a tile in the hardware. Then, it uses an optimization algorithm to place these clusters to the hardware, maximizing performance of the machine learning application on the hardware. Temperature gradients are not incorporated in the mapping process.
    \item \textbf{Proposed:} In this technique the neurons and synapses of an SNN are mapped to the hardware considering the thermal gradient. It uses the clustering technique of SpiNeMap to generate clusters of neurons and synapses, where each cluster can fit on to the resources of a tile. The clusters are mapped to the crossbar using a hill-climbing approach to minimize the average temperature. This reduces the leakage power consumption.
\end{itemize}

\subsection{Evaluated Metrics}
We evaluate the following metrics.
\begin{itemize}
    \item \textbf{Average Temperature:} \mr{This is the average temperature of each crossbar in the hardware. We report the highest average temperatures of all crossbars.}
    \item \textbf{Leakage Power:} This is the total leakage power consumed in the hardware.
    \item \mr{\textbf{Performance:} This is the latency, i.e., the time it takes to execute each model on hardware.}
    \item \textbf{Compilation Time:} This is the time it takes to generate the minimum temperature mapping of an application for the hardware.
\end{itemize}

\section{Results and Discussion}\label{sec:results}
\subsection{Average Temperature}\label{sec:avg_temperature}
\begin{figure}[h!]
	\centering
	\centerline{\includegraphics[width=0.99\columnwidth]{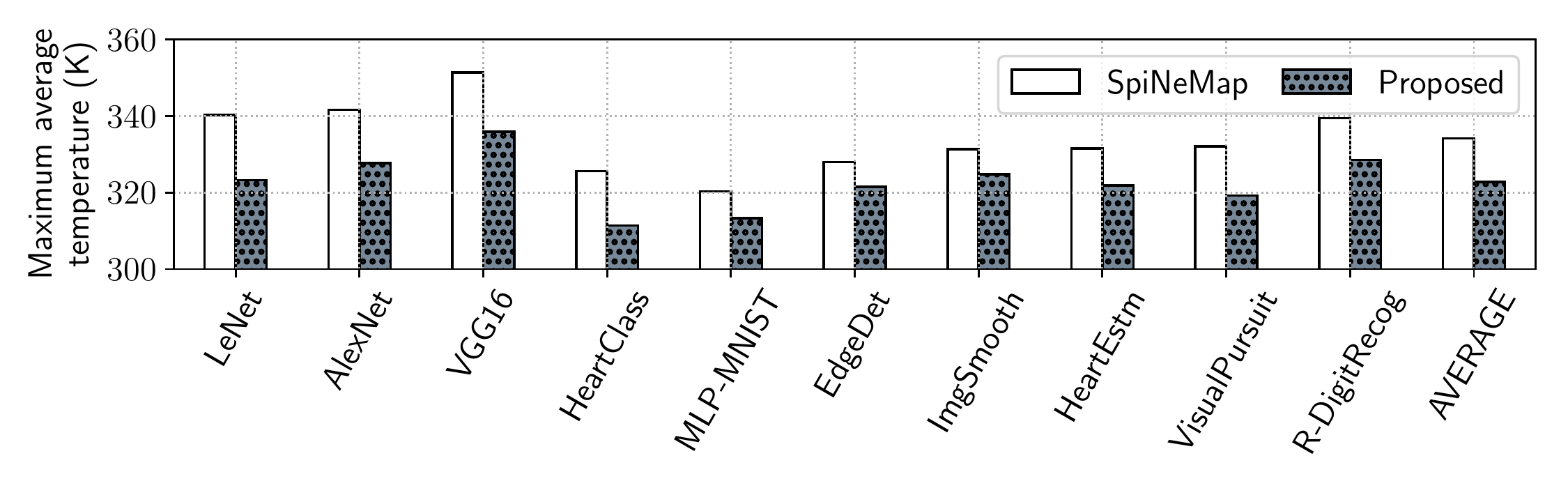}}
	\vspace{-10pt}
	\caption{Maximum average temperature of the crossbars on DYNAP-SE.}
	\vspace{-10pt}
	\label{fig:hardware_average_temperature}
\end{figure}

Figure \ref{fig:hardware_average_temperature} compares the maximum average temperature of the crossbars for each evaluated application on DYNAP-SE using SpiNeMap and the proposed technique.
We make the following \textit{two} key observations.

First, the maximum average temperature increases with model size. VGG16, which has more neurons and synapses than AlexNet (see Table~\ref{tab:apps}), results in higher average temperature than AlexNet for both SpiNeMap and the proposed technique. MLP-MNIST, on the other hand, have lower temperature than both these models due to its lower model complexity. Although R-DigitRecog has comparatively fewer neurons and synapses, the average temperature is much higher. This is because R-DigitRecog has higher activation, i.e., spikes in its workload, which increases the temperature. These results clearly demonstrate the \textit{workload-dependent} nature of the temperature obtained on the hardware. 
Second, the temperature obtained using the proposed mapping technique is lower than SpiNeMap by an average 11.4K (between 6.4K and 17K) for these 10 applications. This reduction is because of the proposed hill climbing algorithm (Algorithm~\ref{alg:min_temp}), which incorporates the thermal gradient in optimizing the mapping of neurons and synapses to the crossbars of the hardware. 

\subsection{Leakage Power}
\begin{figure}[h!]
	\centering
	\vspace{-10pt}
	\centerline{\includegraphics[width=0.99\columnwidth]{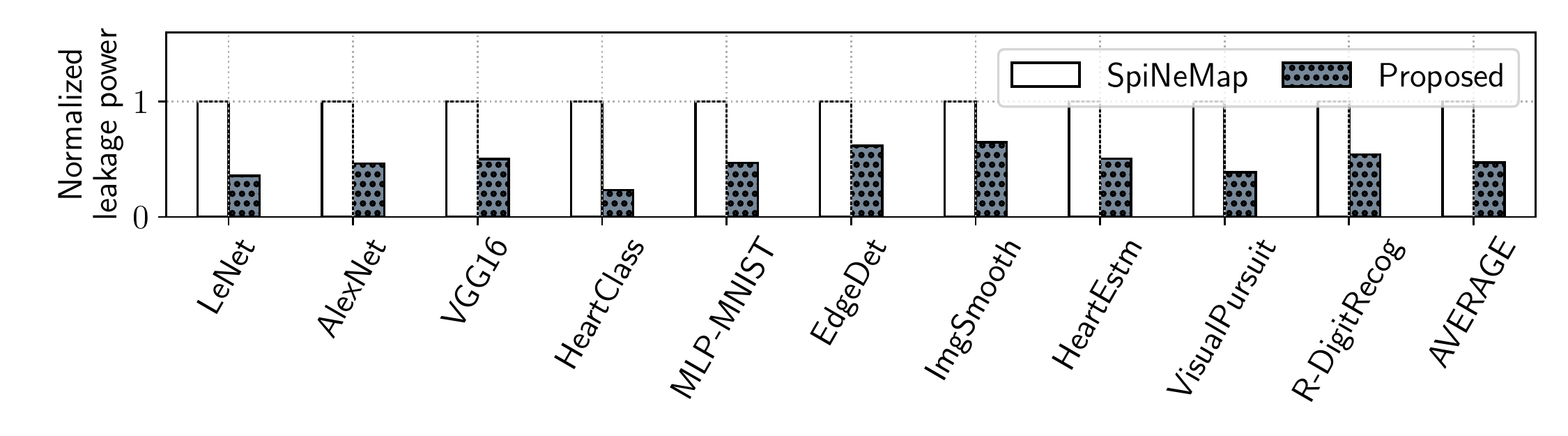}}
	\vspace{-10pt}
	\caption{Normalized leakage power on DYNAP-SE.}
	\vspace{-10pt}
	\label{fig:leakage_power}
\end{figure}

\mr{
Figure \ref{fig:leakage_power} compares the leakage power on DYNAP-SE for each evaluated application using SpiNeMap and the proposed technique.
The leakage power constitute between 20\%--30\% (average 22.8\%) of the total energy consumption in the hardware.
}
Results are normalized with respect to the leakage power obtained on the hardware using SpiNeMap. 
We observe that
the leakage power obtained using the proposed technique is lower than SpiNeMap by an average 52\%. This significant improvement in the leakage power is due to the reduction of the average temperature of the crossbars, which we analyzed in Section~\ref{sec:avg_temperature}. 
\mr{
This reduction in leakage power results in a reduction of the total energy consumption by 11\%.
}

\subsection{\mr{Performance}}
\begin{figure}[h!]
	\centering
	\vspace{-10pt}
	\centerline{\includegraphics[width=0.99\columnwidth]{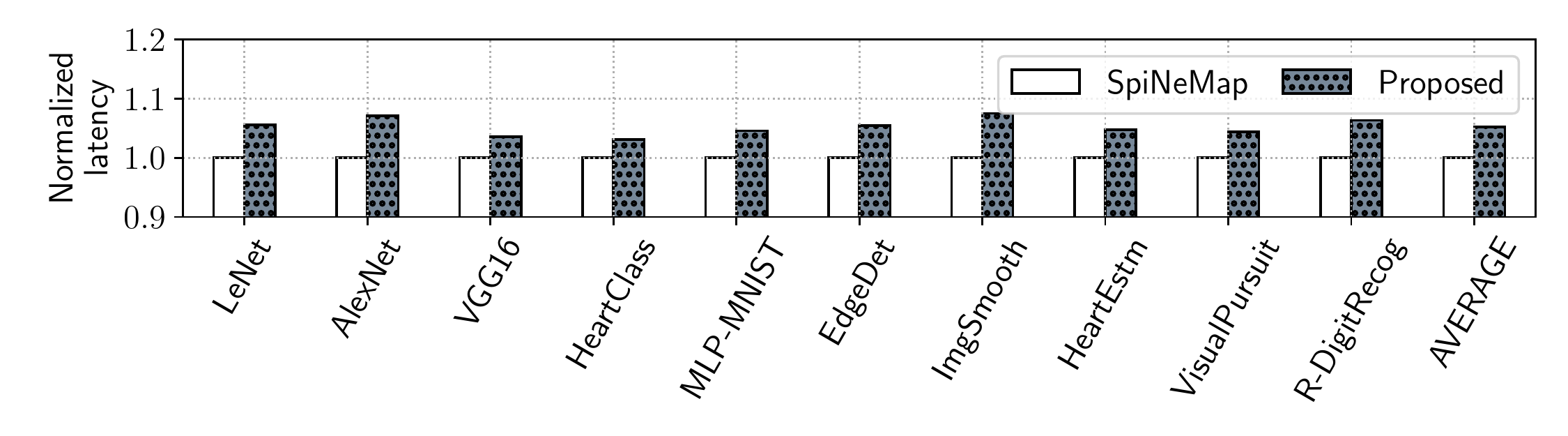}}
	\vspace{-10pt}
	\caption{Normalized latency on DYNAP-SE.}
	\vspace{-10pt}
	\label{fig:performance}
\end{figure}

\mr{Figure~\ref{fig:performance} compares the latency of SpiNeMap and the proposed technique on DYNAP-SE for the evaluated applications. We observe that the latency of the proposed technique is only 5\% higher (average) than SpiNeMap. Although the optimization objective of SpiNeMap (which is performance) is different from the optimization objective of the proposed technique (which is temperature), the proposed technique uses the clustering technique of SpiNeMap to first generate clusters, minimizing the spike communication on the shared interconnect of the hardware. This results in lower spike latency. Therefore, in the next step when the proposed technique optimizes for temperature during placement of the clusters to crossbars of the hardware, the latency is not significantly higher than SpiNeMap.
}

\subsection{Thermal Model Validation}
\mr{
We validate our thermal model against 1) the thermal model of~\cite{xi2011spice}, which models the temperature of a single PCM cell and 2) the detailed model of~\cite{das2015reliability}, which performs a detailed layout-based thermal simulations.
The individual PCM cell model is fast. However, it does not incorporate the thermal contributions from neighboring PCM cells in a crossbar. Therefore, this model is not accurate. On the other hand, the model in~\cite{das2015reliability} is accurate because it incorporates the spatial thermal contributions. However, it takes 30 minutes of wall clock time to perform each thermal simulation for a 128x128 crossbar. Therefore, incorporating this model in Algorithm~\ref{alg:min_temp} to evaluate the temperature of a mapping makes the exploration time infeasible. Instead, we validated our spatial formulation (Equation~\ref{eq:surrounding}) by incorporating this equation into the framework of~\cite{das2015reliability}.
}

\begin{figure}[h!]
	\centering
	\vspace{-10pt}
	\centerline{\includegraphics[width=0.99\columnwidth]{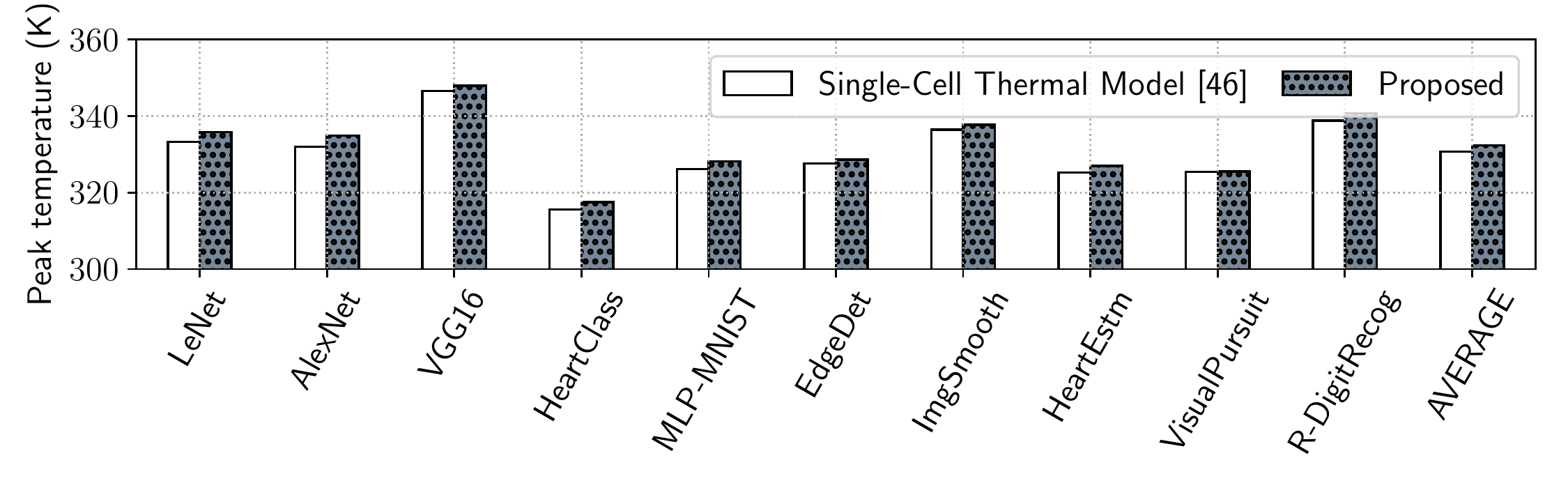}}
	\vspace{-10pt}
	\caption{Comparison of Peak temperature.}
	\vspace{-10pt}
	\label{fig:peak_temperature}
\end{figure}

Figure~\ref{fig:peak_temperature} plots the peak temperature obtained using the model of~\cite{xi2011spice} and the proposed model (Equation~\ref{eq:final_TPCM} \& \ref{eq:surrounding}) for each evaluated application on DYNAP-SE. We observe that existing models such as~\cite{xi2011spice} lead to underestimation of the peak temperature by an average 1.6K for these applications. This is because they do not incorporate the spatial dependency. Underestimation of temperature leads to an underestimation of the leakage power consumption of the hardware.

\begin{figure}[h!]
	\centering
	\vspace{-10pt}
	\centerline{\includegraphics[width=0.99\columnwidth]{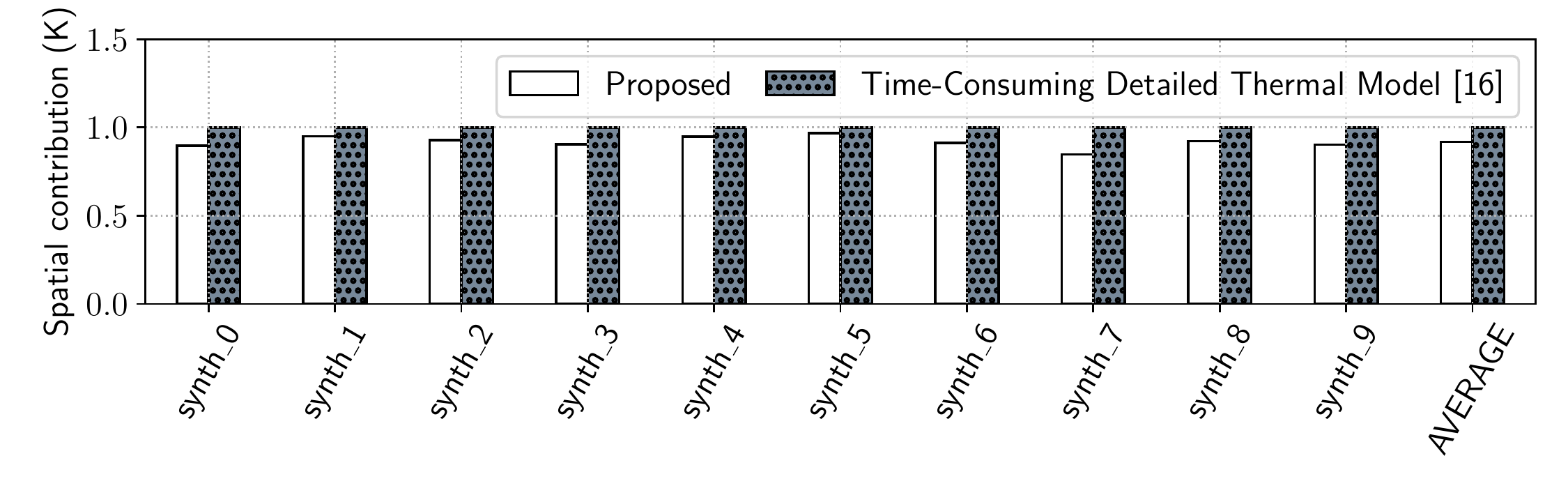}}
	\vspace{-10pt}
	\caption{Comparison of spatial contribution.}
	\vspace{-10pt}
	\label{fig:spatial}
\end{figure}

Figure~\ref{fig:spatial} plots the spatial contribution obtained using the model of~\cite{das2015reliability} and the proposed model (Equation~\ref{eq:surrounding}) for 10 synthetic applications. We observe that the accuracy of the proposed spatial model is close to that of the detailed model~\cite{das2015reliability}. The spatial contribution obtained using Equation~\ref{eq:surrounding} is on average 8.2\% lower than~\cite{das2015reliability} (0.3K in absolute terms).

These results validate the thermal model proposed in this work.

\subsection{Compilation Time and Solution Tradeoff}
Table~\ref{tab:compile_time} reports the compilation time and the average temperature obtained for three different settings of the variable $MaxIter$. We observe that as $MaxIter$ is increased, the average temperature reduces for all applications. This is because with increase in the number of iterations, Algorithm~\ref{alg:min_temp} is able to find a better solution. However, the compilation time also increases. Finally, we observe that increasing $MaxIter$ from 100 to 1000 results in a significant increase in compilation time with a minimal improvement of the average temperature. We conclude that setting $MaxIter = 100$ gives the best trade-off in terms of compilation time and the solution quality.
\mr{
User can use this $MaxIter$ parameter to set a limit on the compilation time of their algorithm by analyzing the complexity of their model against the ones we evaluate (see Table~\ref{tab:apps}).
}

\vspace{-10pt}

\begin{table}[h!]
	\renewcommand{\arraystretch}{1.2}
	\setlength{\tabcolsep}{1.2pt}
	\caption{Compilation time and solution tradeoff.}
	\label{tab:compile_time}
	\centering
	{\fontsize{6}{9}\selectfont
		\begin{tabular}{|r|c|c|c|c|c|c|}
			\hline
			\multirow{4}{*}{\textbf{Application}} & \multicolumn{2}{|c|}{$\mathbf{MaxIter = 10}$} & \multicolumn{2}{|c|}{$\mathbf{MaxIter = 100}$} & \multicolumn{2}{|c|}{$\mathbf{MaxIter = 1000}$}\\
			\cline{2-7}
			& \textbf{Compilation} & \textbf{Avg.} & \textbf{Compilation} & \textbf{Avg.} & \textbf{Compilation} & \textbf{Avg.}\\
			& \textbf{Time} & \textbf{Temperature} & \textbf{Time} & \textbf{Temperature} & \textbf{Time} & \textbf{Temperature}\\
			& \textbf{(sec)} & \textbf{(K)} & \textbf{(sec)} & \textbf{(K)} & \textbf{(sec)} & \textbf{(K)}\\
			\hline
			LeNet & 26 & 326.3 & 259 & 323.2 & 2641	& 322.2\\
            AlexNet & 114 & 330.1 & 1144 & 327.6 & 11480 & 326.0\\
            VGG16 & 241 & 344.6 & 2413 & 335.8 & 24180 & 335.3\\
            HeartClass & 96 & 315.1 & 965 & 311.3 & 9699 & 309.9\\
            MLP-MNIST & 14 & 319.7 & 149 & 313.2 & 1520 & 311.6\\
            EdgeDet & 12 & 323.5 & 132 & 321.5 & 1337 & 320.8\\
            ImgSmooth & 26 & 327.11 & 268 & 324.7 & 2740 & 322.8\\
            HeartEstm & 12 & 328.2 & 125 & 321.8 & 1255 & 320.4\\
            VisualPursuit & 27 & 329.1 & 284	 & 319.2 & 2883 & 318.7\\
            R-DigitRecog & 15 & 336.3 & 159 & 328.5 & 1615 & 327.9\\
    \hline
	\end{tabular}}
\end{table}

\section{Conclusions}\label{sec:conclusions}
We propose a technique to map the neurons and synapses of SNN-based machine learning applications to neuromorphic hardware. Prior work in this space have focused extensively on performance, with no consideration of the thermal aspects and the associated leakage power problem in the hardware. Our technique is based on two key contributions. First, we propose a new thermal model of a crossbar incorporating contributions from the adjacent cells. Second, we incorporate this thermal model in a hill-climbing approach to minimize the average temperature across the crossbars of the hardware.  We evaluate our approach using 10 machine learning applications and show the significant reduction of the average temperature of the hardware.
By lowering the average temperature, we also show a reduction of leakage power consumption.

\vspace{-10pt}

\section{Acknowledgment}
This work is supported by the National Science Foundation Faculty Early Career Development Award CCF-1942697 (CAREER: Facilitating Dependable Neuromorphic Computing: Vision, Architecture, and Impact on Programmability).

\vspace{-10pt}

\bibliographystyle{splncs04}
\bibliography{disco,external}
%




\end{document}